\pdfoutput=1
\documentclass[11pt]{article}

\usepackage{emnlp2021}

\usepackage{times}
\usepackage{latexsym}

\usepackage[T1]{fontenc}
\usepackage[utf8]{inputenc}

\usepackage{microtype}

\usepackage[utf8]{inputenc}
\usepackage{graphicx}
\usepackage{amsmath}
\usepackage{amsthm}
\usepackage{booktabs}
\usepackage{algorithm}
\usepackage{algorithmic}
\usepackage{amssymb}
\usepackage{multirow}
\usepackage{array}
\usepackage{setspace}
\usepackage{pifont}
\usepackage{adjustbox}
\usepackage{hyperref}
\usepackage{makecell}

\title{Improving Distantly-Supervised Named Entity Recognition with Self-Collaborative Denoising Learning}

\author{Xinghua Zhang, Bowen Yu, Tingwen Liu$^{\ast}$\thanks{$^{\ast}$Corresponding author}, Zhenyu Zhang,\\ {\bf Jiawei Sheng,} {\bf Mengge Xue} \and {\bf Hongbo Xu} \\
        Institute of Information Engineering, Chinese Academy of Sciences, Beijing, China \\ 
        School of Cyber Security, University of Chinese Academy of Sciences, Beijing, China \\
        \texttt{\{zhangxinghua,yubowen,liutingwen,zhangzhenyu1996\}@iie.ac.cn}\\
        \texttt{\{shengjiawei,xuemengge,hbxu\}@iie.ac.cn}
        }

\makeatletter
\def\thanks#1{\protected@xdef\@thanks{\@thanks
        \protect\footnotetext{#1}}}
\makeatother
\begin{document}
\maketitle

\begin{abstract}
Distantly supervised named entity recognition (DS-NER) efficiently reduces labor costs but meanwhile intrinsically suffers from the label noise due to the strong assumption of distant supervision.
Typically, the wrongly labeled instances comprise numbers of incomplete and inaccurate annotation noise,
while most prior denoising works are only concerned with one kind of noise and fail to fully explore useful information in the whole training set. To address this issue, we propose a robust learning paradigm named Self-Collaborative Denoising Learning (SCDL), which jointly trains two teacher-student networks in a mutually-beneficial manner to iteratively perform noisy label refinery. Each network is designed to exploit reliable labels via self denoising, and two networks communicate with each other to explore unreliable annotations by collaborative denoising. Extensive experimental results on five real-world datasets demonstrate that SCDL is superior to state-of-the-art DS-NER denoising methods\footnote{The source code and data can be found at \url{https://github.com/AIRobotZhang/SCDL}.}. 
\end{abstract}

\section{Introduction}

Named Entity Recognition (NER) is the task of detecting entity spans and then classifying them into predefined categories, such as person, location and organization. Due to the capability of extracting entity information and benefiting many NLP applications (e.g., relation extraction~\cite{lin-et-al:attention}, question answering~\cite{li-et-al:answering}), NER appeals to many researchers. Traditional supervised methods for NER require a large amount of high-quality corpus for model training, which is extremely expensive and time-consuming as NER requires token-level labels.

\begin{figure}[tp]
\centering
\includegraphics[width=7.8cm]{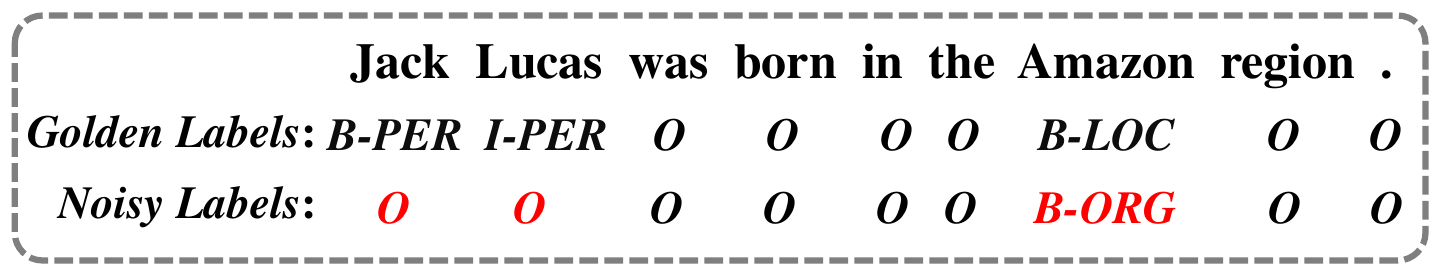}
\caption{A noisy sample generated by distantly-supervised methods, where {\it Jack Lucas} is the incomplete annotation and {\it Amazon} is inaccurate.}
\label{Figure 1}
\end{figure}

Therefore, in recent years, distantly supervised named entity recognition (DS-NER) has been proposed to automatically generate labeled training set by aligning entities in knowledge bases (e.g., Freebase) or gazetteers to corresponding entity mentions in sentences. This labeling procedure is based on a strong assumption that each entity mention in a sentence is a positive instance of the corresponding type according to the extra resources. However, this assumption is far from reality. Due to the limited coverage of existing resources, many entity mentions in the text cannot be matched and are wrongly annotated as non-entity, resulting in incomplete annotations. Moreover, two entity mentions with the same surface name can belong to different entity types, thus simple matching rules may fall into the dilemma of labeling ambiguity and produce inaccurate annotations. As illustrated in Figure~\ref{Figure 1}, the entity mention ``{\it Jack Lucas}'' is not recognized due to the limited coverage of extra resources and ``{\it Amazon}'' is wrongly labeled with organization type owing to the labeling ambiguity.

Recently, many denoising methods~\cite{shang-et-al:dictionary,yang-et-al:learning,cao-et-al:data,peng-et-al:learning,li-et-al:recognition} have been developed to handle noisy labels in DS-NER. %Some CRF-variant models~\cite{shang-et-al:dictionary,yang-et-al:learning,cao-et-al:data} were proposed to avoid fitting noisy labels. ~\citeauthor{shang-et-al:dictionary}~\shortcite{shang-et-al:dictionary} also proposed a new tagging scheme (AutoNER) to tolerate the label noise. 
For example, ~\citeauthor{shang-et-al:dictionary}~\shortcite{shang-et-al:dictionary} obtained high-quality phrases through {\it AutoPhrase}~\cite{shang-et-al:corpora} and designed AutoNER to model these phrases that may be potential entities. ~\citeauthor{peng-et-al:learning}~\shortcite{peng-et-al:learning} proposed a positive-unlabeled learning algorithm to unbiasedly and consistently estimate the NER task loss, and ~\citeauthor{li-et-al:recognition}~\shortcite{li-et-al:recognition} used negative sampling to eliminate the misguidance brought by unlabeled entities. Though achieving good performance, most studies mainly focus on solving incomplete annotations with a strong assumption of no inaccurate ones existing in DS-NER. Meanwhile, these methods aim to reduce the negative effect of noisy labels by weakening or abandoning the wrongly labeled instances. Hence, they can at most alleviate the noisy supervision and fail to fully mine useful information from the mislabeled data. Intuitively, if we can rectify those unreliable annotations into positive instances for model training, a higher data utilization and better performance will be achieved. We argue that an ideal DS-NER denoising system should be capable of solving two kinds of label noise (i.e., incomplete and inaccurate annotations) and making full use of the whole training set. %Unfortunately, going with any of the existing approaches, it's difficult to make the best of both worlds.

In this work, we strive to reconcile this gap and propose a robust learning framework named SCDL (Self-Collaborative Denoising Learning). SCDL co-trains two teacher-student networks to form inner and outer loops for coping with label noise without any assumption, as well as making full exploration of mislabeled data. The inner loop inside each teacher-student network is a self denoising scheme to select reliable annotations from two kinds of noisy labels, and the outer loop between two networks is a collaborative denoising procedure to rectify unreliable instances into useful ones. Specifically, in the inner loop, each teacher-student network selects consistent and high-confidence labeled tokens generated by the teacher to train the student, and then updates the teacher gradually via exponential moving average (EMA)\footnote{A momentum technique that has been explored in several studies, e.g., Adam~\cite{kingma-ba:optimization}, semi-supervised~\cite{tarvainen-valpola:results} and self-supervised~\cite{grill-et-al:learning} learning.} based on the re-trained student. And as for the outer loop, the high-quality pseudo labels generated by one network’s teacher are used to update the noisy labels of the other network thanks to the stability of EMA and different noise sensitivities between two networks. Moreover, the inner and outer loop procedures will be performed alternately. Obviously, a successful self denoising process (inner loop) can generate high-quality pseudo labels which benefit the collaborative learning procedure (outer loop) a lot and a promising outer loop will promote the inner loop by refining noisy labels, thus handling the label noise in DS-NER effectively. % and making full use of the training set.
%thus freeing the trade-off dilemma when handling incomplete and inaccurate annotations simultaneously.

We evaluate our method on five DS-NER datasets. Experimental results indicate that SCDL consistently achieves superior performance over previous competing approaches. Extensive validation studies demonstrate the rationality and robustness of our self-collaborative denoising framework.

\section{Related Work}

Many studies have obtained reliable performance in NER. For example, BiLSTM-CRF~\cite{lample-et-al:recognition} and BERT~\cite{devlin-et-al:understanding} based methods become the paradigm in NER due to their promising performances. However, most of these works rely on high-quality labels, which are quite expensive. To address this issue, several studies attempted to annotate tokens via distant supervision~\cite{liang-et-al:supervision}. They matched unlabeled sentences with external gazetteers or knowledge Graphs (KGs). Despite the success of distant supervision, it still suffers from noisy labels (i.e., incomplete and inaccurate annotations in NER).

\paragraph{DS-NER Denoising.} Many studies~\cite{shang-et-al:dictionary,cao-et-al:data,jie-et-al:recognition} tried to modify the standard CRF for adapting to the scenario of label noise, e.g., Fuzzy CRF. ~\citeauthor{ni-et-al:projection}~\shortcite{ni-et-al:projection} selected high-confidence labeled data from noisy data to train NER models. And many new training paradigms were proposed to resist label noise in DS-NER, such as AutoNER~\cite{shang-et-al:dictionary}, Reinforcement Learning~\cite{yang-et-al:learning,nooralahzadeh-et-al:annotation}, AdaPU~\cite{peng-et-al:learning} and Negative Sampling~\cite{li-et-al:recognition}. In addition, some studies~\cite{mayhew-et-al:data,liang-et-al:supervision} performed iterative training procedures to mitigate noisy labels in DS-NER. However, most studies mainly focus on incomplete annotations regardless of inaccurate ones or depending on manually labeled data. What's more, most prior methods are insufficient since they can at most alleviate the negative effect caused by label noise and fail to mine useful information from the whole training set. Different from previous studies, we propose two denoising learning procedures which can be enhanced each other mutually with the devised teacher-student network and co-training paradigm, mitigating two kinds of label noise and making full use of the whole training set.
%Thus, inaccurate and incomplete annotation noises can be reduced effectively. Several studies~\cite{yang-et-al:learning,cao-et-al:data} have proposed to handle both inaccurate and incomplete annotation noises,

\paragraph{Teacher-Student Network.} The teacher-student network is well known in knowledge distillation~\cite{hinton-et-al:network}. A teacher is generally a complicated model and the light weight student imitates its output. Recently, there are many variations of teacher-student network. For example, self-training copies the student as a new teacher to generate pseudo labels~\cite{xie-et-al:classification,wang-et-al:detection}. ~\citeauthor{liang-et-al:supervision}~\shortcite{liang-et-al:supervision} applied self-training with teacher-student network to handle label noise in DS-NER. However, for the teacher-student network in our framework, the teacher selects reliable annotations with devised strategies for training student and then we use EMA to update the teacher based on re-trained student. With this loop, our method can learn entity knowledge effectively.

\paragraph{Co-Training.} The co-training paradigm which jointly trains two models is used to improve the robustness of models~\cite{blum-mitchell:training,nigam-ghani:training,kiritchenko-matwin:training}. Many previous frameworks~\cite{han-et-al:labels,yu-et-al:corruption,wei-et-al:regularization,li-et-al:learning} have adopted co-training to denoise, but they mainly use the diversity of two single models and the single one doesn't have the denoising ability. But supervision signals from the peer model are not always clean. Instead, we train two groups of teacher-student networks and each group can also perform label denoising effectively which further improves the co-training paradigm.

\section{Task Definition}

Given the training corpus $\mathcal{D}$ where each sample is a form of ($X_i$, $Y_i$), $X_i=x_1, x_2, ..., x_N$ represents a sentence with {\it N} tokens and $Y_i=y_1, y_2, ..., y_N$ is the corresponding tag sequence. Each entity mention $e=x_i, ..., x_j (0 \le i \le j \le N)$ is a span of the text , associated with an entity type, e.g., person, location. In this paper, we use the BIO scheme following~\cite{liang-et-al:supervision}. In detail, the begin token of an entity mention is labeled as {\it B-type} and others are {\it I-type}. The non-entity tokens are annotated as {\it O}.

The traditional NER problem is a supervised learning task by fitting a sequence labeling model based on the training dataset. However, we mainly explore the practical scenario when the labels of training data are contaminated due to the distant supervision. In other words, the revealed tag $y_i$ may not correspond to the underlying correct one. The challenge posed in this setting is to reduce the negative influence of noisy annotations and generate high-confidence labels for them to make full use of the training data.

\section{Methodology}

\begin{figure*}[tp]
\centering
\includegraphics[scale=0.25]{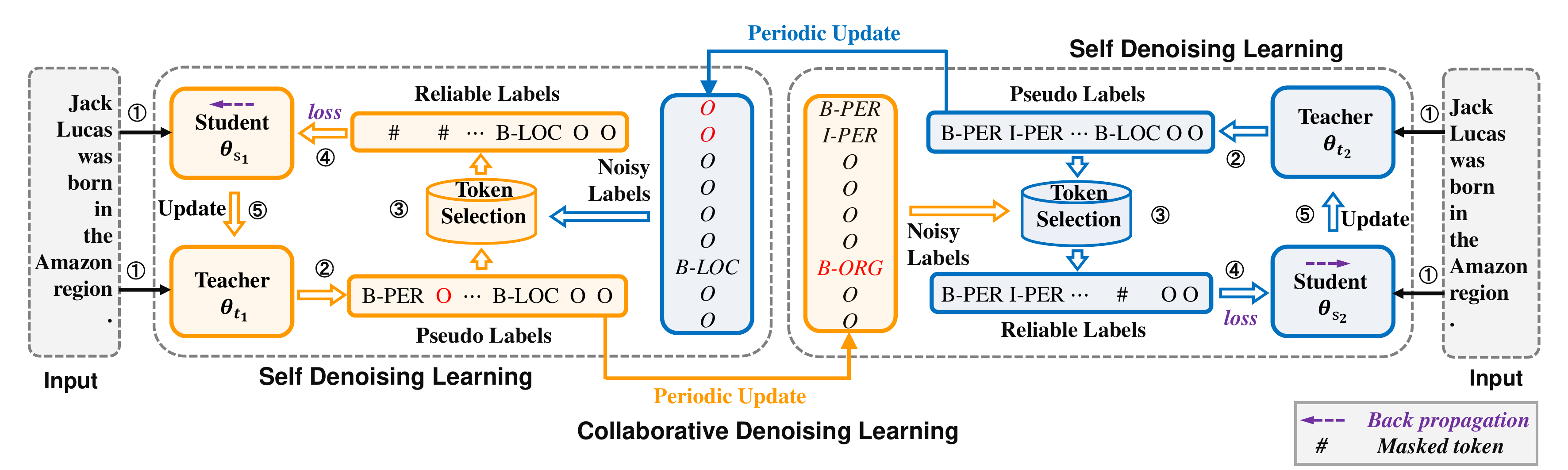}
\caption{Overview of SCDL with two procedures performed iteratively. (1) Each teacher-student network contributes to an \textbf{inner loop} (i.e., \textbf{self denoising}): [\ding{173}] the teacher first generates pseudo labels to [\ding{174}] select tokens along with noisy labels, then [\ding{175}] the student is optimized based on selected tokens, and finally [\ding{176}] the teacher is updated by the student. (2) The interplay between two teacher-student networks is an \textbf{outer loop} (i.e., \textbf{collaborative denoising}): the pseudo labels are applied to update the noisy labels of the peer network periodically.}
\label{Figure 2}
\end{figure*}

In this section, we give a detailed description of our self-collaborative denoising learning framework, which consists of two interactive teacher-student networks to address both the incomplete and inaccurate annotation issues. As illustrated in Figure~\ref{Figure 2}, each teacher-student network contributes to an inner loop for self denoising and the outer loop between two networks is a collaborative denoising scheme. These two procedures can be optimized in a mutually-beneficial manner, thus improving the performance of the NER system.

\subsection{Self Denoising Learning}

It is widely known that deep neural networks have high capacity for memorization~\cite{arpit-et-al:networks}. When noisy labels become prominent, deep neural NER models inevitably overfit noisy labeled data, resulting in poor performance. The purpose of self denoising learning is to select reliable labels to reduce the negative influence of noisy annotations. To achieve this end, self denoising learning involves a teacher-student network, where the teacher first generates pseudo labels to participate in labeled token selection, then the student is optimized via back-propagation based on selected tokens, and finally the teacher is updated by gradually shifting the weights of the student in continuous training with exponential moving average (EMA). We take two neural NER models with the same architecture as the teacher and student respectively.

\subsubsection{Labeled Token Selection}
This subsection illustrates our labeled token selection strategy based on the consistency and high confidence predictions.
\paragraph{Consistency Predictions.} It has been observed that the model’s predictions of wrongly labeled instances fluctuate drastically in previous studies~\cite{huang-et-al:networks}. A mislabeled instance will be supervised by both its wrong label and similar instances. For example, {\it Amazon} is wrongly annotated as {\it organization} in Figure~\ref{Figure 1}. The wrong label {\it organization} pushes the model to fit this supervision signal while other clean tokens with similar context will encourage the model to predict it as {\it location}. Therefore, we can take advantage of this property to separate clean tokens from noisy ones.

Based on above analysis, how to quantify the fluctuation becomes a key issue. One straightforward solution is to integrate predictions from different training iterations but with more time-space complexity. Thanks to the widespread concern of EMA, we use it to update the teacher’s parameters. In this way, the teacher can be viewed as the temporal ensembling of the student models in different training steps and then its prediction will be the ensemble of predictions from past iterations. Therefore, the pseudo labels predicted by the teacher can quantify the fluctuation of noisy labels naturally. Subsequently, we devise the first token selection strategy based on the fluctuation of noisy labels to identify the correctly labeled tokens $(\bar{X}_i, \bar{Y}_i)$ via the consistency between noisy labels and predicted pseudo labels, denoted as:
\begin{align}
    {(\bar{X}_i, \bar{Y}_i)}_{\rm{CP}}=\{(x_j, y_j)~|~ y_j=\tilde{y}_j, \tilde{y}_j \in f(X_i; \theta_t)\}
    % L=\{(x_j, y_j)~|~ y_j=\tilde{y}_j, y_j \in Y_i, \tilde{y}_j \in f(X_i; \theta_t)\}
\end{align}
where $y_j \in Y_i$ is the noisy label of the {\it j}-th token in the {\it i}-th sentence and $\tilde{y}_j$ is the pseudo label predicted by the teacher $\theta_t$.

\paragraph{High Confidence Predictions.} As studied in previous works~\cite{bengio-et-al:learning,arpit-et-al:networks}, hard samples can not be learnt effectively at first, thus predictions of those mislabeled hard samples may not fluctuate and then they are mistakenly believed to be reliable. To alleviate this issue, we propose the second selection strategy to pick tokens with high confidence predictions, as formulated in Equation~\ref{Equation 2}, where $\tilde{p}_j$ is the label distribution of the {\it j}-th token predicted by the teacher, $\delta$ denotes the confidence threshold.
\begin{align}
    {(\bar{X}_i, \bar{Y}_i)}_{\rm{HCP}}=\{(x_j, y_j)~|~ {\rm{max}}(\tilde{p}_j) \ge \delta\}
\label{Equation 2}
\end{align}

\subsubsection{Optimization}
\paragraph{Loss Function of the Student.} Standard supervised NER methods are fitting the outputs of a model to hard labels (i.e, one-hot vectors) to optimize the parameters. However, when the model is trained with tokens and mismatched hard labels, wrong information is being provided to the model. Compared with hard labels, the supervision with soft labels is more robust to the noise because it carries the uncertainty of the predicted results. Therefore, we modify the standard cross entropy loss into a soft label form defined as:
\begin{align}
    \mathcal{L}(\theta_s)=-\frac{1}{MN}\sum_{i=1}^M \sum_{j=1}^N \sum_{c=1}^C \mathbb{I}_{i,j}\tilde{p}_{j,c}^i~ {\rm{log}}(p_{j,c}^i)
\end{align}
\begin{align}
    \mathcal{T}_i={(\bar{X}_i, \bar{Y}_i)}_{\rm{CP}} \cap {(\bar{X}_i, \bar{Y}_i)}_{\rm{HCP}}
\end{align}%
where $p_{j,c}^i$ is the probability of the {\it j}-th token with the {\it c}-th class in the {\it i}-th sentence predicted by the student and $\tilde{p}_{j,c}^i$ is from the teacher. $\mathcal{T}_i$ includes the tokens in the {\it i}-th sentence meeting the consistency and high confidence selection strategies simultaneously. $\mathbb{I}$ is the indicator function, $\mathbb{I}_{i,j}=1$ when the {\it j}-th token is in $\mathcal{T}_i$, otherwise $\mathbb{I}_{i,j}$ is 0.

Then the parameters of the student model can be updated via back-propagation as follows:
\begin{align}
    \theta_s \gets \theta_s - \gamma \frac{\partial  \mathcal{L}}{\partial \theta_s}
\end{align}

\paragraph{Update of the Teacher.} Different from the optimization of the student model, we apply EMA to gradually update the parameters of the teacher, as shown in Equation~\ref{Equation 6}, where $\alpha$ denotes the smoothing coefficient.
\begin{align}
    \theta_t \gets \alpha \theta_t + (1-\alpha) \theta_s
\label{Equation 6}
\end{align}
Although the clean token selection strategies indeed alleviate noisy annotations, they also suffer from unreliable token choice which misguides the model into generating biased predictions. As formulated in Equation~\ref{Equation 7}, the update of the teacher $\theta^i_t$ in {\it i}-th iteration can be converted into the form of back-propagation (derivations in Appendix A.1):
\begin{align}
    \theta^i_t&=\theta^{i-1}_t - \gamma (1-\alpha) \sum_{j=0}^{i-1}\alpha^{i-1-j} \frac{\partial \mathcal{L}}{\partial \theta^{j}_s}
\label{Equation 7}
\end{align}
where $\gamma$ is the learning rate and $(1-\alpha)$ is a small number because $\alpha$ is generally assigned a value close to 1 (e.g., 0.995), equivalent to multiplying a small coefficient on the weighted sum of student’s past gradients. Therefore, with the conservative and ensemble property, the application of EMA has largely mitigated the bias. As a result, the teacher tends to generate more reliable pseudo labels, which can be used as new supervision signals in the collaborative denoising phase.

\begin{table*}[tb]
\renewcommand\arraystretch{1.2}
\centering
\resizebox{15.9cm}{3cm}{
\begin{tabular}{llccccccccccccccc}
\toprule
\multicolumn{2}{c}{\multirow{2}{*}{\textbf{Method}}} & \multicolumn{3}{c}{\textbf{CoNLL03}} & \multicolumn{3}{c}{\textbf{OntoNotes5.0}} & \multicolumn{3}{c}{\textbf{Webpage}} & \multicolumn{3}{c}{\textbf{Wikigold}} & \multicolumn{3}{c}{\textbf{Twitter}} \\ \cline{3-17} 
\multicolumn{2}{l}{} & \textbf{P} & \textbf{R} & \textbf{F1} & \textbf{P} & \textbf{R} & \textbf{F1} & \textbf{P} & \textbf{R} & \textbf{F1} & \textbf{P} & \textbf{R} & \textbf{F1} & \textbf{P} & \textbf{R} & \textbf{F1} \\ \midrule
\multirow{2}{*}{(\romannumeral1)}
& $\textbf{BiLSTM-CRF}^{\clubsuit}$  & 91.35 & 91.06 & 91.21 & 85.99 & 86.36 & 86.17 & 50.07 & 54.76 & 52.34 & 55.40  & 54.30  & 54.90  & 60.01 & 46.16 & 52.18 \\
& $\textbf{RoBERTa}^{\clubsuit}$  & 89.14 & 91.10  & 90.11 & 84.59 &87.88 & 86.20  & 66.29 &79.73 & 72.39 &85.33 & 87.56 & 86.43 & 51.76 & 52.63 & 52.19 \\ \midrule
\multirow{4}{*}{(\romannumeral2)}
& \textbf{KB-Matching}  & 81.13 & 63.75 & 71.40 & 63.86 & 55.71 & 59.51 & 62.59 &45.14 & 52.45 & 47.90 & 47.63 & 47.76 & 40.34 & 32.22 & 35.83 \\
& $\textbf{BiLSTM-CRF}^{\dag}$ & 75.50 & 49.10 & 59.50 & \textbf{68.44} & 64.50 & 66.41 & 58.05 & 34.59 & 43.34 & 47.55 & 39.11 & 42.92 & 46.91 & 14.18 & 21.77 \\
& $\textbf{DistilRoBERTa}^{\dag \star}$ & 77.87 & 69.91 & 73.68 & 66.83 & 68.81 & 67.80 & 56.05  & 59.46  & 57.70  & 48.85  & 52.05 & 50.40 & 45.72 & 43.85 & 44.77 \\
& $\textbf{RoBERTa}^{\dag \star}$ & 82.29       & 70.47       & 75.93       & 66.99       & 69.51       & 68.23       & 59.24       & 62.84       & 60.98       & 47.67       & 58.59       & 52.57       & 50.97       & 42.66       & 46.45      \\ \midrule
\multirow{6}{*}{(\romannumeral3)}
& $\textbf{AutoNER}^{\ddag}$~\cite{shang-et-al:dictionary} & 75.21       & 60.40        & 67.00          & 64.63       & 69.95           & 67.18       & 48.82       & 54.23       & 51.39       & 43.54       & 52.35       & 47.54       & 43.26       & 18.69       & 26.10       \\
& $\textbf{LRNT}^{\ddag}$~\cite{cao-et-al:data}    & 79.91       & 61.87       & 69.74       & 67.36       & 68.02       & 67.69       & 46.70        & 48.83       & 47.74       & 45.60        & 46.84       & 46.21       & 46.94       & 15.98       & 23.84      \\
& $\textbf{Co-teaching+}^{\ddag \star}$~\cite{yu-et-al:corruption}          & 86.04       & 68.74       & 76.42       & 66.63       & 69.32       & 67.95       & 61.65       & 55.41       & 58.36       & 55.23       & 49.26       & 52.08       & 51.67       & 42.66       & 46.73      \\
& $\textbf{JoCoR}^{\ddag \star}$~\cite{wei-et-al:regularization}   & 83.65       & 69.69       & 76.04       & 66.74       & 68.74       & 67.73       & 62.14       & 58.78       & 60.42       & 51.48       & 51.23       & 51.35       & 49.40        & \textbf{45.59}       & 47.42      \\
& $\textbf{NegSampling}^{\ddag \star}$~\cite{li-et-al:recognition}    & 80.17       & 77.72             & 78.93       & 64.59       & \textbf{72.39}       & 68.26        & \textbf{70.16}       & 58.78       & 63.97       & 49.49       & 55.35             & 52.26       & 50.25       & 44.95             & 47.45 \\
& $\textbf{BOND}^{\ddag}$~\cite{liang-et-al:supervision}    & 82.05       & \textbf{80.92}             & 81.48       & 67.14       & 69.61       & 68.35       & 67.37       & 64.19       & 65.74       & 53.44       & \textbf{68.58}             & 60.07       & 53.16       & 43.76             & 48.01       \\ \midrule
& \textbf{SCDL (Ours)}             & \textbf{87.96}             & 79.82       & \textbf{83.69}             & 67.49       & 69.77       & \textbf{68.61}             & 68.71             & \textbf{68.24}             & \textbf{68.47}             & \textbf{62.25}             & 66.12       & \textbf{64.13}             & \textbf{59.87}             & 44.57       & \textbf{51.09}            \\ \bottomrule 
\end{tabular}}
\caption{Main results on five benchmark datasets. (\romannumeral1) $\clubsuit$ marks the model trained on the fully clean dataset. (\romannumeral2) $\dag$ marks the model trained on noisy dataset without label denoising. (\romannumeral3) $\ddag$ marks the prior label denoising framework. $\star$ marks produced with official implementation.}
\label{Table 1}
\end{table*}

\subsection{Collaborative Denoising Learning}

Based on the devised clean token selection strategy in self denoising learning, the teacher-student network can utilize the correctly labeled tokens in an ideal situation to alleviate the negative effect of label noise. However, just filtering unreliable labeled tokens will inevitably lose useful information in training set since there is no opportunity for the wrongly labeled tokens to be corrected and explored. Intuitively, if we can change the wrong label to the correct one, it will be transformed into a useful training instance.

Inspired by some co-training paradigms~\cite{han-et-al:labels,yu-et-al:corruption,wei-et-al:regularization}, we propose the collaborative denoising learning to update noisy labels mutually for mining more useful information from dataset by deploying two teacher-student networks with different architecture. As stated in~\cite{bengio:minima}, a human brain can learn more effectively if guided by the signals produced by other humans. Similarly, the pseudo labels predicted by the teacher are applied to update the noisy labels of the peer teacher-student network periodically since two teacher-student networks have different learning abilities based on different initial conditions and network structures. With this outer loop, the noisy labels can be improved continuously and the training set can be fully explored.

% In fact, training two models has been widely applied in computer vision (e.g., Co-teaching~\cite{han-et-al:labels}, Co-teaching+~\cite{yu-et-al:corruption}, JoCoR~\cite{wei-et-al:regularization}) to improve the robustness of trained models. In this paper, we propose to train two teacher-student networks for improving label noise in NER. What's more, we train two groups of networks instead of two single ones and each group of teacher-student has the ability to denoise which is not available in the former approaches based on training two models.

% \begin{tiny}
\begin{algorithm}[tb]
% \fontsize{6pt}{1pt}
\small
\setstretch{1.15}
\captionsetup{font={small}}
\caption{Training Procedure of SCDL}
\label{alg:algorithm}
\textbf{Input}: Training corpus  $\mathcal{D}=\{(X_i, Y_i)\}_{i=1}^M$ with noisy labels \\
\textbf{Parameter}: Two network parameters $\theta_{t_1}$, $\theta_{s_1}$, $\theta_{t_2}$, and $\theta_{s_2}$ \\
\textbf{Output}: The best model
\begin{algorithmic}[1] %[1] enables line numbers
\STATE Pre-training two models $\theta_1$, $\theta_2$ with $\mathcal{D}$. \hfill $\triangleright${\it Pre-Training}.\\
\STATE $\theta_{t_1} \gets \theta_1$, $\theta_{s_1} \gets \theta_1$, $\theta_{t_2} \gets \theta_2$, $\theta_{s_2} \gets \theta_2$, $step \gets 0$.\\
% \FOR {$p \gets 1\ to\ P$}
\STATE Initialize noisy labels: $Y_{\uppercase\expandafter{\romannumeral1}} \gets Y, Y_{\uppercase\expandafter{\romannumeral2}} \gets Y$.\\
\WHILE{ \emph{not reach max training epochs} }
    \STATE Get a batch $(X^{(b)}, Y^{(b)}_{\uppercase\expandafter{\romannumeral1}}, Y^{(b)}_{\uppercase\expandafter{\romannumeral2}})$ from $\mathcal{D}$, \\
    $step \gets step+1$. \hfill $\triangleright${\it Self Denoising Learning}.\\
    \STATE Get pseudo-labels via the teacher $\theta_{t_1}$, $\theta_{t_2}$: \\ $ \tilde{Y}^{(b)}_{\uppercase\expandafter{\romannumeral1}} \gets f(X^{(b)}; \theta_{t_1}) $, \\ $ \tilde{Y}^{(b)}_{\uppercase\expandafter{\romannumeral2}} \gets f(X^{(b)}; \theta_{t_2})$. \\
    \STATE Get clean tokens: \\ $\mathcal{T}^{(b)}_{\uppercase\expandafter{\romannumeral1}} \gets$ TokenSelection$(Y^{(b)}_{\uppercase\expandafter{\romannumeral1}},  \tilde{Y}^{(b)}_{\uppercase\expandafter{\romannumeral1}})$, \\ $\mathcal{T}^{(b)}_{\uppercase\expandafter{\romannumeral2}} \gets$ TokenSelection$(Y^{(b)}_{\uppercase\expandafter{\romannumeral2}},  \tilde{Y}^{(b)}_{\uppercase\expandafter{\romannumeral2}})$. \\
    \STATE Update the student $\theta_{s_1}$ and $\theta_{s_2}$ by Eq. 3 and Eq. 5. \\
    \STATE Update the teacher $\theta_{t_1}$ and $\theta_{t_2}$ by Eq. 6. \\
    \IF { $step \text{ mod } Update\_Cycle = 0$ }
        \STATE Update noisy labels mutually: \hfill $\triangleright${\it Collaborative Denoising Learning}.\\
        $Y_{\uppercase\expandafter{\romannumeral1}}=\{Y_i \gets f(X_i; \theta_{t_2})\}_{i=1}^M$, \\
        $Y_{\uppercase\expandafter{\romannumeral2}}=\{Y_i \gets f(X_i; \theta_{t_1})\}_{i=1}^M$.

    \ENDIF
\ENDWHILE
\STATE Evaluate models $\theta_{t_1}$, $\theta_{s_1}$, $\theta_{t_2}$, $\theta_{s_2}$ on {\it Dev} set. \\
\STATE \textbf{return} The best model $\theta\in\{ \theta_{t_1}, \theta_{s_1}, \theta_{t_2}, \theta_{s_2}\}$
\end{algorithmic}
\end{algorithm}

\subsection{Algorithm Workflow}

In this subsection, we introduce the overall procedure of our SCDL framework. Algorithm 1 gives the pseudocode. To summarize, the training process of SCDL can be divided into three procedures: (1) \textbf{Pre-Training with Noisy Labels.} We warm up two NER models $\theta_1$ and $\theta_2$ on the noisy labels to obtain a better initialization, and then duplicate the parameters $\theta$ for both the teacher $\theta_t$ and the student $\theta_s$ (i.e., $\theta_{t_1}$= $\theta_{s_1}$= $\theta_1$, $\theta_{t_2}$= $\theta_{s_2}$= $\theta_2$). The training objective function in this stage is the cross entropy loss with the following form:
\begin{align}
    \mathcal{L}(\theta)=-\frac{1}{MN}\sum_{i=1}^M \sum_{j=1}^N y_j^i {\rm{log}}(p(y_j^i|X_i;\theta))
\end{align}
where $y_j^i$ means the {\it j}-th token label of the {\it i}-th sentence in the
noisy training corpus and $p(y_j^i|X_i;\theta)$ denotes its probability produced by model $\theta$. {\it M} and {\it N} are the size of training corpus and the length of sentence respectively. (2) \textbf{Self Denoising Learning.} In this stage, we can select correctly labeled tokens to train the two teacher-student networks respectively. (3) \textbf{Collaborative Denoising Learning.} Self denoising can only utilize correct annotations and this phase will update noisy labels mutually to relabel tokens for two teacher-student networks. The initial noisy labels of two networks comes from distant supervision. The second and third phase are conducted alternately, which will promote each other to perform label denoising. It’s worth noting that only the best model $\theta\in\{ \theta_{t_1}, \theta_{s_1}, \theta_{t_2}, \theta_{s_2}\}$ is adopted for predicting.

\section{Experiments}

In this section, we evaluate the performance of SCDL, compared with several comparable baselines. Additionally, we conduct lots of auxiliary experiments and provide comprehensive analyses to justify the effectiveness of SCDL.

\subsection{Experimental Settings}
\paragraph{Datasets.} We conduct experiments on five publicly available NER datasets: \textbf{CoNLL03}~\cite{sang-meulder:recognition}, \textbf{OntoNotes5.0}~\cite{weischedel-et-al:ldc2013t19}, \textbf{Webpage}~\cite{ratinov-roth:recognition}, \textbf{Wikigold}~\cite{balasuriya-et-al:wikipedia} and \textbf{Twitter}~\cite{godin-et-al:representations}. ~\citeauthor{liang-et-al:supervision}~\shortcite{liang-et-al:supervision} re-annotated the training set by distant supervision, and left the development and test set unchanged. The statistics of datasets are in Appendix A.2.

\paragraph{Baselines and Evaluation Metrics.} We compare our method with several competitive baselines from three aspects. (\romannumeral1) {\it Fully-Clean.} \textbf{BiLSTM-CRF}~\cite{ma-hovy:crf} and \textbf{RoBERTa}~\cite{liu-et-al:approach} are fully trained on clean dataset (without noisy labels) for NER, as the upper bound of denoising. (\romannumeral2) {\it Fully-Noisy.} \textbf{KB-Matching} uses distant supervision to annotate test set. \textbf{BiLSTM-CRF}, \textbf{DistilRoBERTa} and \textbf{RoBERTa} are trained on noisy dataset without label denoising, as the lower bound of denoising. (\romannumeral3) {\it Label-Denoising.} We compare several DS-NER denoising baselines which propose to solve noisy labels. \textbf{AutoNER}~\cite{shang-et-al:dictionary} and \textbf{LRNT}~\cite{cao-et-al:data} try to reduce the negative effect of noisy labels, leaving training dataset unexplored fully. \textbf{Co-teaching+}~\cite{yu-et-al:corruption} and \textbf{JoCoR}~\cite{wei-et-al:regularization} are two classical label denoising methods, developed in computer vision. \textbf{NegSampling}~\cite{li-et-al:recognition} only handles incomplete annotations by negative sampling. \textbf{BOND}~\cite{liang-et-al:supervision} adapts self-training directly to DS-NER, suffering from confirmation bias (a problem from self-training itself). We use Precision (\textbf{P}), Recall (\textbf{R}) and \textbf{F1} score as the evaluation metrics.
% including \textbf{KB-Matching}, \textbf{BiLSTM-CRF}~\cite{ma-hovy:crf}, \textbf{RoBERTa}~\cite{liu-et-al:approach}, \textbf{AutoNER}~\cite{shang-et-al:dictionary}, \textbf{LRNT}~\cite{cao-et-al:data}, \textbf{BOND}~\cite{liang-et-al:supervision}, \textbf{NegSampling}~\cite{li-et-al:recognition} and classical denoising methods \textbf{Co-teaching+}~\cite{yu-et-al:corruption} and \textbf{JoCoR}~\cite{wei-et-al:regularization}. We use Precision (\textbf{P}), Recall (\textbf{R}) and \textbf{F1} score as the evaluation metrics.
% \paragraph{Baselines and Evaluation Metrics.} We compare our method with several baseline models, including \textbf{KB-Matching}, \textbf{BiLSTM-CRF}~\cite{ma-hovy:crf}, \textbf{RoBERTa}~\cite{liu-et-al:approach}, \textbf{AutoNER}~\cite{shang-et-al:dictionary}, \textbf{LRNT}~\cite{cao-et-al:data}, \textbf{BOND}~\cite{liang-et-al:supervision}, \textbf{NegSampling}~\cite{li-et-al:recognition} and classical denoising methods \textbf{Co-teaching+}~\cite{yu-et-al:corruption} and \textbf{JoCoR}~\cite{wei-et-al:regularization}. We use Precision (\textbf{P}), Recall (\textbf{R}) and \textbf{F1} score as the evaluation metrics.

\paragraph{Implementation Details.} For fair comparison, we adopt RoBERTa ($\theta_1$) and DistilRoBERTa ($\theta_2$) as the basic models. The max training epochs is 50, and the confidence threshold $\delta$ is 0.9. The batch size is set to 16 or 32, the learning rate is 1e-5 or 2e-5 according to different datasets. We tune EMA parameter $\alpha$ from \{0.9,0.99,0.995,0.998\}, tune update cycle according to the size of dataset (e.g., 6000 iterations (about 7 epochs) for CoNLL03) on development set. We implement our code with Pytorch based on huggingface Transformers\footnote{https://huggingface.co/transformers/}. Detailed hyperparameter settings for each dataset and tuning procedures are listed in Appendix A.3.

\subsection{Experimental Results}

Table~\ref{Table 1} shows the results of our proposed method compared with baselines and highlights the best overall performance in bold. Obviously, SCDL achieves the best performance, and improves the precision as well as F1 score significantly, compared with previous state-of-the-art models.

Compared to our basic models (i.e., DistilRoBERTa and RoBERTa), SCDL improves the F1 score with an average increase of 8.33\% and 6.37\% respectively, which demonstrates the necessity of label denoising in the distantly-supervised NER task and the effectiveness of the proposed method.

In addition, SCDL performs much better than previous studies which consider the noisy labels in NER, including AutoNER, LRNT, NegSampling and BOND. The reason is that they mainly focus on one kind of label noise in DS-NER or fail to make full use of the mislabeled data with their strategies. On contrast, our method can not only exploit correctly labeled tokens but also explore valuable information in wrongly labeled ones by correction. Compared to the popular denoising methods in computer vision: Co-teaching+ and JoCoR, SCDL gains of up to 12.05\% absolute percentage points in F1 score. We guess this is beacause most computer vision denoising studies focus on instance-level classification, while NER is a token-level task where non-entity category accounts for the majority, and this case is not fully considered. Thus corruption occurs easily in DS-NER denoising task for these methods as the training goes.

\begin{table}[tb]
\renewcommand\arraystretch{1.2}
\centering
% \tiny
\resizebox{6.7cm}{1.62cm}{
\begin{tabular}{lccc}
\toprule
 & \textbf{P}     & \textbf{R}     & \textbf{F1}    \\ \midrule
\textbf{SCDL}            & 89.42 & 80.74 & \textbf{84.86} \\ \midrule
w/o consistency prediction  & 87.01  & \textbf{81.11} & 83.96        \\
w/o high confidence prediction  & 88.14  & 80.94  & 84.38       \\
w/o $\theta_{t_2}$, $\theta_{s_2}$ (co-training paradigm)      & 88.45  & 78.32 & 83.08       \\
w/o $\theta_{t_1}$, $\theta_{t_2}$ (teacher-student network)       & 87.90  & 77.22  & 82.22        \\
w/o soft labels & \textbf{89.86}  & 79.12  & 84.15        \\ \bottomrule
\end{tabular}}
\caption{Ablation study on CoNLL03 dev set.}
\label{Table 2}
\end{table}

\subsection{Analysis}
\paragraph{Ablation Study.} To evaluate the influence of each component in our method, we conduct the ablation study for further exploration (see Table~\ref{Table 2}). Overall, although SCDL is not optimal on precision or recall, it achieves the best in F1 score, which indicates that our method can balance well when taking two kinds of annotation noise into account and exploring full training set. Based on these ablations, we observe that: (1) Token selection strategy with the consistency and high confidence predictions indeed promote the overall performance (F1 score) by improving the precision and marginally lowering the recall. The recall value doesn’t decrease sharply in our framework because of the unbiased predictions generated by teacher model and alternate optimization. (2) When we keep only one teacher-student network (i.e., w/o $\theta_{t_2}$, $\theta_{s_2}$), both recall and F1 decrease visibly, which validates the effectiveness of collaborative denoising learning since more wrongly labeled tokens (e.g., false negative tokens) can be explored via the peer dynamic update of noisy labels. (3) Meanwhile, removing two teacher models (i.e., w/o $\theta_{t_1}$, $\theta_{t_2}$) leads to the decline on both precision and recall. Because this simplification impairs the devised teacher-student network. It uses the predictions of each student to support the token selection strategies and the mutual update of noisy labels, which loses the stable optimization ability of EMA and leads to unreliable token selection. (4) Learning from noisy annotations benefits from soft labels since they contain the uncertainty of predicted results and are more tolerant to the noise compared to the hard ones.

\begin{figure}[tp]
\centering
\includegraphics[width=8cm]{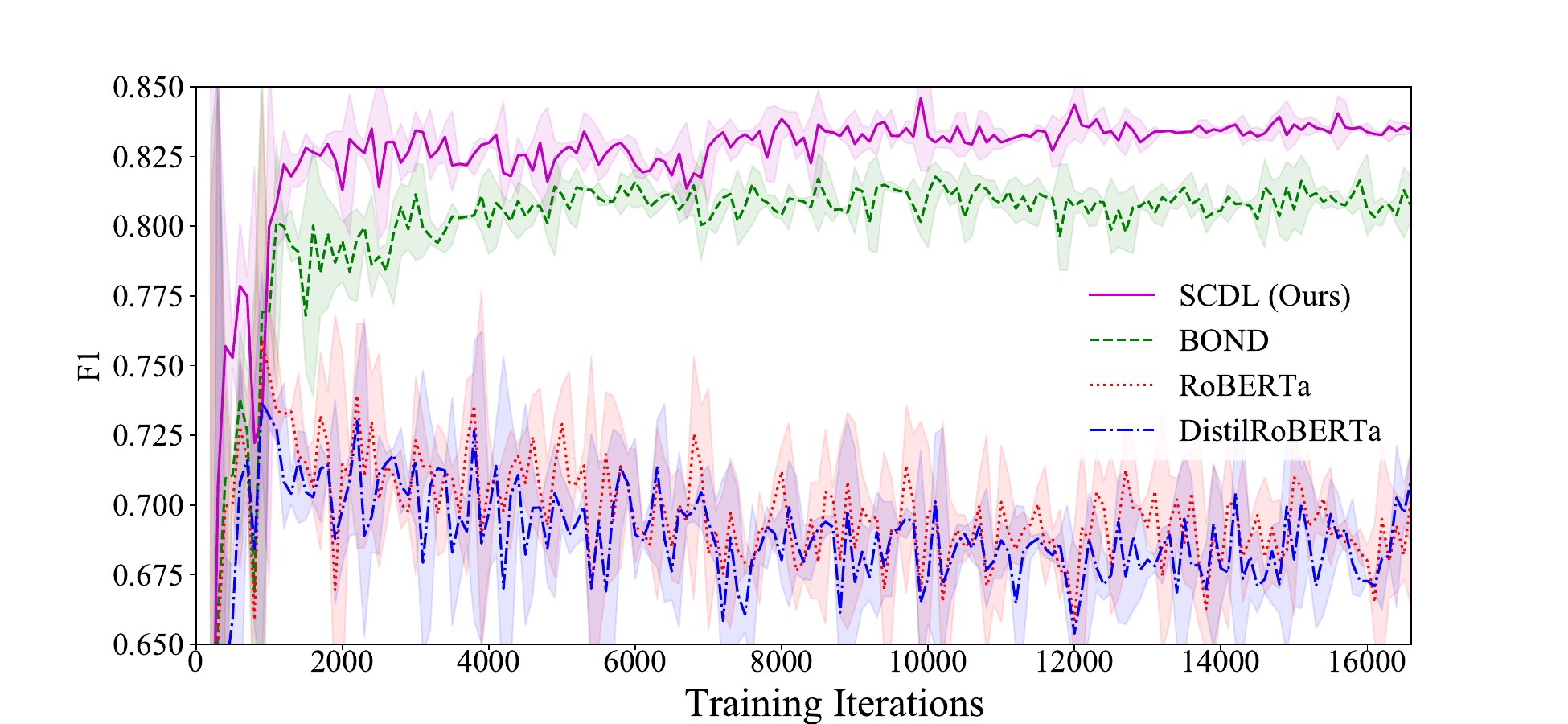}
\caption{Learning curves of SCDL and other baselines about F1 score vs. training iterations on CoNLL03.}
\label{Figure 3}
\end{figure}
\paragraph{Learning Curve of SCDL.} To evaluate the advantage of the proposed framework in handling noisy labels during training, we show the F1 score vs. training iterations on CoNLL03 test set in Figure~\ref{Figure 3}. Compared to RoBERTa and DistilRoBERTa, the performance of SCDL and BOND remains stable as the training goes. Because of the memorization effect of networks, the F1 score of RoBERTa and DistilRoBERTa first reach a high level and then gradually decrease. Moreover, SCDL consistently achieves better performance than other baselines at almost any training stage, which again confirms the effectiveness of our denoising framework.

\begin{figure*}[tbp]
\centering
\includegraphics[scale=0.27]{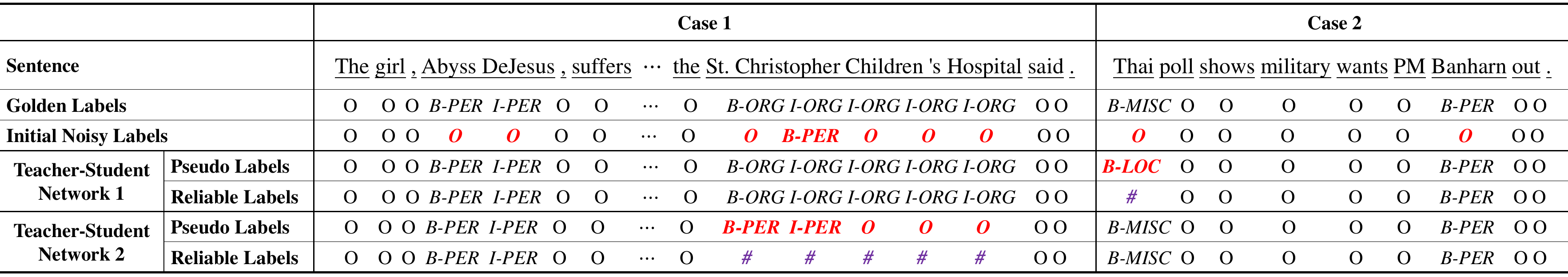}
\caption*{Table 4: Case studies. Wrong labels are marked in red and \# means the masked token (i.e., not selected).}
\end{figure*}

\begin{figure}[!tbp]
\centering
\includegraphics[width=8cm]{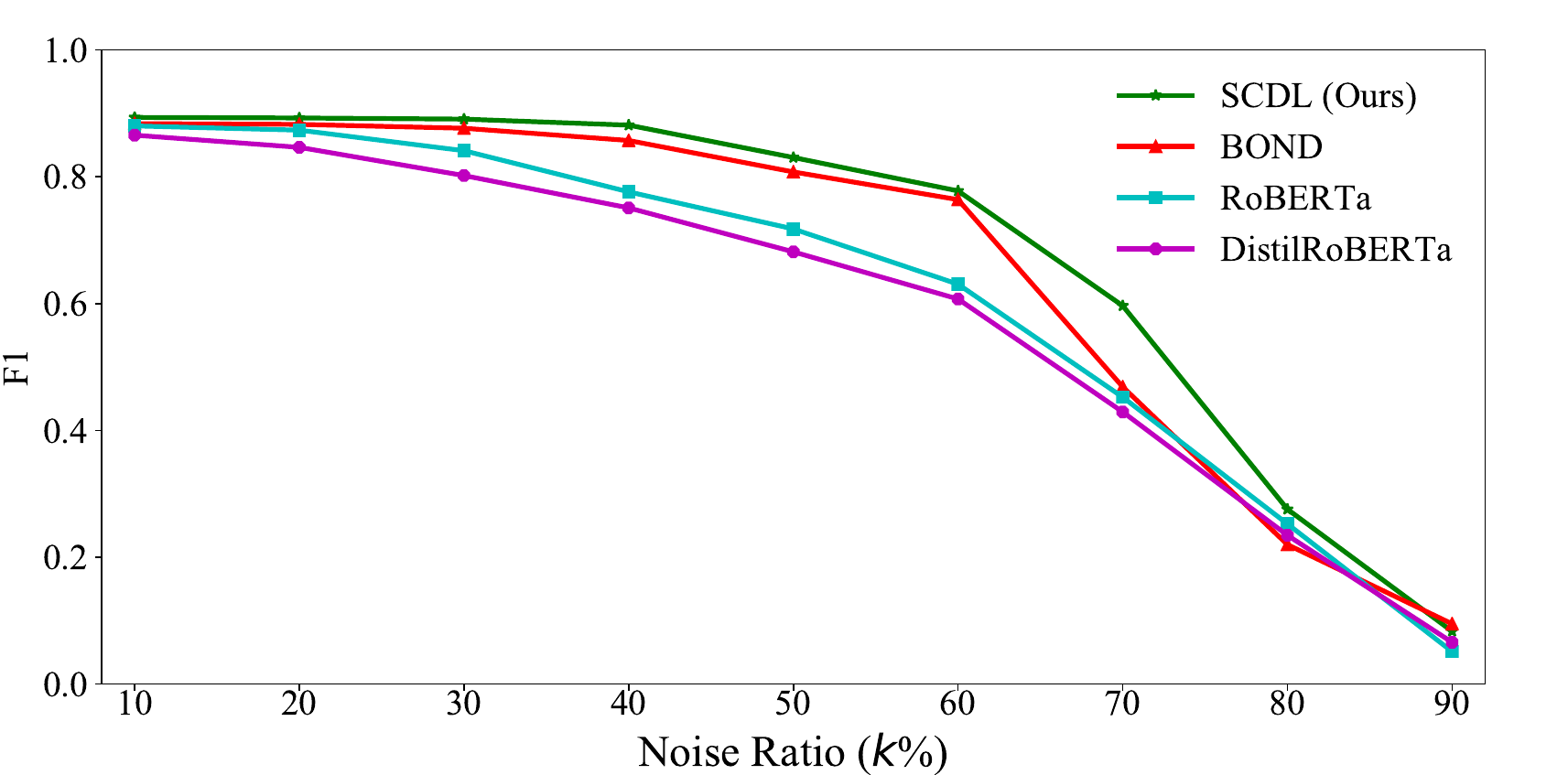}
\caption{F1 on CoNLL03 with different noise ratio.}
\label{Figure 4}
\end{figure}
\paragraph{Robustness to Different Noise Ratio.} To study the robustness of the proposed method in different noise ratio, we randomly replace $k\%$ entity mention labels in the corpus with other entity types or non-entity to construct different proportions of label noise and report the test F1 score on CoNLL03 in Figure~\ref{Figure 4}. The pre-trained language models (e.g., RoBERTa) are robust to low level noise (less than 20\%) due to their strong expressive power. When the noise ratio is between 30\% and 80\%, SCDL is more robust and exhibits satisfactory denoising ability, since the training data still has reasonable entity type knowledge and SCDL can learn from it to refine noisy labels. However, both SCDL and BOND degenerate to the basic model in the hardest case (more than 80\%) which may not exist in reality and needs further studies in the future.

\paragraph{Effectiveness of Noisy Label Refinery.} As the noisy labels are updated dynamically during training to explore the full dataset, we compare the F1 score before and after denoising on training set, as shown in Table~\ref{Table 3}. In detail, SCDL refines noisy labels on CoNLL03 and Twitter training set, from 70.97 to 81.22, 37.73 to 50.82 respectively, which surpasses BOND. The reason may be that BOND mainly depends on self-training which suffers from confirmation bias, while SCDL can bypass this issue by the devised teacher-student network and co-training paradigm and then improves both precision and recall significantly. Overall, the comparison before and after denoising demonstrates that SCDL indeed refines the training noisy labels to a certain extent, leading to the better use of the mislabeled data and outstanding performance on test.

\begin{table}[t]
\renewcommand\arraystretch{1.2}
\centering
% \tiny
\resizebox{7.0cm}{1.12cm}{
\begin{tabular}{lcccccc}
\toprule
\multirow{2}{*}{} & \multicolumn{3}{c}{\textbf{CoNLL03}} & \multicolumn{3}{c}{\textbf{Twitter}} \\ \cline{2-7} 
& \textbf{P} & \textbf{R} & \textbf{F1} & \textbf{P} & \textbf{R} & \textbf{F1} \\ \midrule
\textbf{Distant-Supervision}   & 82.38 & 62.33 & 70.97 & 46.71 & 31.64 & 37.73 \\
\textbf{BOND-Denoising}   & 80.42  & \textbf{76.46}  & 78.39  & 53.76 & 34.82 & 42.27 \\
\textbf{SCDL-Denoising}   & \textbf{87.42}  & 75.85   & \textbf{81.22} & \textbf{54.86}  & \textbf{47.33} & \textbf{50.82}   \\ \bottomrule
\end{tabular}}
\caption{Comparison of denoising ability of SCDL and BOND on training set.}
\label{Table 3}
\end{table}
%We focus on inaccurate and incomplete annotation noises in this paper. SCDL is proposed to handle them better simultaneously, where each group of teacher-student network selects accurate labeled tokens to learn more reliable entity knowledge and co-training two groups of them corrects more false negative entity mentions to make full use of training data. 
%Previous denoising studies (e.g., BOND~\cite{liang-et-al:supervision}) also involve teacher-student network in self-training, but they only copy the student as the new teacher to relabel tokens. Instead, we use EMA to update the teacher based on re-trained student and the teacher selects reliable annotations with devised strategies for training student. With this loop, SCDL can exploit the detected accurate samples to learn more effectively. Though co-training paradigm has been widely applied in denoising (e.g., JoCoR~\cite{wei-et-al:regularization}), they only co-train two single models to improve the robustness of single model to noisy labels. While we co-train two groups of networks to explore correcting unreliable samples, and each group also has the ability of denoising. Because the exploration of outer loop is not always reliable. Thus SCDL alleviates two kinds of noise in NER with the help of exploit and exploration in self and collaborative denoising learning. Table 4 gives an intuitive case study to justify the effectiveness of SCDL. More examples can be found in Appendix D.
\paragraph{Case Study.} Different from most prior denoising studies on DS-NER, our proposed framework SCDL can not only handle two kinds of label noise (i.e., inaccurate and incomplete annotations) without any assumption, but also make full use of the whole training set. High F1 score in Table~\ref{Table 1} and the effectiveness of noisy label refinery in Table~\ref{Table 3} have proved the feasibility of SCDL quantitatively. For better understanding intuitively, we give two samples from CoNLL03 after two periodic updates to show the denoising ability of SCDL in Table 4. For case 1 with two kinds of label noise, the person name ``{\it Abyss DeJesus}'' and organization name ``{\it St. Christopher Children 's Hospital}'' are not correctly annotated by DS-NER. After denoising, ``{\it Abyss DeJesus}'' is corrected and transformed into a useful instance. Though the hospital name is still not corrected in the {\it teacher-student network 2}, but {\it network 2} selects reliable annotations successfully for training student. It shows that SCDL can not only exploit reliable instances but also explore unreliable ones. Similar situations also occur in case 2, while the {\it network 2} has better capability which demonstrates the validity of co-training paradigm. %More examples can be found in Appendix D.

\paragraph{Efficiency Analysis.} In training stage, with the same batch size, the serial efficiency of our method is about 1.5 batches per second on single GPU Tesla T4, other baselines like BOND is 2.6, Co-teaching+ is 1.8, JoCoR is 1.9. The memory usage of our method is equivalent to Co-training models (e.g., Co-teaching+). Although there are two student and two teacher models in our method, only two students need back-propagation which occupies the main computational overhead (time and memory usage), while two teachers updated with EMA only need forward-propagation which occupies less computational overhead. It’s worth noting that the two teacher-student networks in our framework can be trained in parallel, which will further accelerate the training. What’s more, compared with other baselines, the test efficiency of our method is the same because we only use one model for predicting.

% \paragraph{Teacher Student Network.} Researchers have made great efforts in named entity recognition (NER), which obtain reliable performance. For example, BiLSTM-CRF~\cite{lample-et-al:recognition} and BERT~\cite{devlin-et-al:understanding} based methods become the paradigm in NER due to their promising performances. However, most of these works rely on high-quality labels, which are quite expensive. To address this issue, several studies attempted to annotate tokens via distant supervision~\cite{yang-et-al:learning,peng-et-al:learning}. They matched unlabeled sentences with external gazetteers or knowledge Graphs (KGs) by string matching, regular expressions or heuristic rules.

% \paragraph{Co-training Paradigm.} Researchers have made great efforts in named entity recognition (NER), which obtain reliable performance. For example, BiLSTM-CRF~\cite{lample-et-al:recognition} and BERT~\cite{devlin-et-al:understanding} based methods become the paradigm in NER due to their promising performances. However, most of these works rely on high-quality labels, which are quite expensive. To address this issue, several studies attempted to annotate tokens via distant supervision~\cite{yang-et-al:learning,peng-et-al:learning}. They matched unlabeled sentences with external gazetteers or knowledge Graphs (KGs) by string matching, regular expressions or heuristic rules.

\section{Conclusion and Future Work}
This paper proposes SCDL to handle two kinds of label noise in DS-NER without any assumption. With devised teacher-student network and co-training paradigm, SCDL can not only exploit more reliable annotations to avoid the negative effect of noisy labels but also explore more useful information from the mislabeled data. Experimental results confirm its effectiveness and robustness in dealing with the label noise. For future work, data augmentation is worth exploring in our framework. Besides, SCDL can also be adapted to other NLP denoising tasks, e.g., classification and matching.
% This paper proposes SCDL to handle two kinds of label noise in DS-NER without any assumption. Each teacher-student network performs self denoising learning to exploit more reliable annotations with devised token selection strategies and the two networks update noisy labels mutually to explore more useful information from training set via collaborative denoising. Moreover, these two procedures can be optimized in a mutually-beneficial manner. Experimental results demonstrate its effectiveness and robustness in alleviating the label noise. For future work, different model initialization and data augmentations for the two teacher-student networks are worth exploring. Additionally, the proposed framework can also be adapted to alleviate the noisy labels in other tasks, such as text classification and matching.

\section*{Acknowledgements}

We would like to thank the anonymous reviewers for their insightful comments and constructive suggestions. This research is supported by the National Key Research and Development Program of China (grant No.2016YFB0801003) and the Strategic Priority Research Program of Chinese Academy of Sciences (grant No.XDC02040400).

\bibliography{anthology,custom}
\bibliographystyle{acl_natbib}

% \appendix

% \section{Example Appendix}
% \label{sec:appendix}

% This is an appendix.
\appendix
\section{Appendix}
\subsection{Derivation of EMA Update}
In this appendix, we give detailed derivation of reorganizing exponential moving average (EMA) as the form of backpropagation. The student $\theta_s$ optimized via back-propagation in the {\it i}-th iteration is shown in Equation~\ref{Equation A-1}, and Equation~\ref{Equation A-2} represents the update process of the teacher $\theta_t$ with EMA.
\begin{align}
    \theta^i_s&=\theta^{i-1}_s - \gamma \frac{\partial \mathcal{L}}{\partial \theta^{i-1}_s}
\label{Equation A-1}
\end{align}

\begin{align}
    \theta^i_t&=\alpha \theta^{i-1}_t + (1-\alpha) \theta^{i}_s
\label{Equation A-2}
\end{align}
Based on Equation~\ref{Equation A-1} and Equation~\ref{Equation A-2}, the teacher $\theta_t$ in the {\it i}-th iteration can be represented as follows:

\begin{align}
  \theta^i_t&=\alpha \theta^{i-1}_t + (1-\alpha) \theta^{i}_s \nonumber\\
  &=\alpha^{i}\theta^{0}_t + \alpha^{i-1}(1-\alpha)(\theta^{0}_s-\gamma \frac{\partial \mathcal{L}}{\partial \theta^{0}_s}) + ... + \nonumber\\
  &+(1-\alpha)(\theta^{0}_s-\gamma \frac{\partial \mathcal{L}}{\partial \theta^{0}_s}-...-\gamma \frac{\partial \mathcal{L}}{\partial \theta^{i-1}_s}) \nonumber\\
  &=\alpha^{i}\theta^{0}_t + (1-\alpha) \sum_{j=0}^{i-1}\alpha^{j}\theta^{0}_s - \gamma (1-\alpha)( \nonumber\\
  &\sum_{j=0}^{i-1}\alpha^{j}\frac{\partial \mathcal{L}}{\partial \theta^{0}_s}
  +\sum_{j=0}^{i-2}\alpha^{j}\frac{\partial \mathcal{L}}{\partial \theta^{1}_s} + ... + \sum_{j=0}^{0}\alpha^{j}\frac{\partial \mathcal{L}}{\partial \theta^{i-1}_s}) \nonumber\\
  &=\alpha^{i}\theta^{0}_t + (1-\alpha)\frac{1-\alpha^{i}}{1-\alpha}\theta^{0}_s - \gamma(1-\alpha)(\nonumber\\
  &\frac{1-\alpha^{i}}{1-\alpha}\frac{\partial \mathcal{L}}{\partial \theta^{0}_s} +
  \frac{1-\alpha^{i-1}}{1-\alpha}\frac{\partial \mathcal{L}}{\partial \theta^{1}_s} + ... + \frac{\partial \mathcal{L}}{\partial \theta^{i-1}_s}) \nonumber\\
  &=\alpha^{i}\theta^{0}_t + \theta^{0}_s - \alpha^{i}\theta^{0}_s - \gamma[(1-\alpha^{i})\frac{\partial \mathcal{L}}{\partial \theta^{0}_s}+ \nonumber\\
  &+ (1-\alpha^{i-1}) \frac{\partial \mathcal{L}}{\partial \theta^{1}_s}+...+(1-\alpha)\frac{\partial \mathcal{L}}{\partial \theta^{i-1}_s}] \nonumber\\
  &=\theta^{0}_s - \gamma\sum_{j=0}^{i-1}(1-\alpha^{i-j})\frac{\partial \mathcal{L}}{\partial \theta^{j}_s} \nonumber\\
  &=\bar{\theta} - \gamma\sum_{j=0}^{i-1}(1-\alpha^{i-j})\frac{\partial \mathcal{L}}{\partial \theta^{j}_s}  \nonumber\\
  &w.r.t. \quad \theta^{0}_s=\theta^{0}_t=\bar{\theta}
    % x =& \prod_{i=1}^n \sum_{j=1}^n j_i + \prod_{i=1}^n \sum_{j=1}^n i_j + \prod_{i=1}^n \sum_{j=1}^n j_i + \prod_{i=1}^n \sum_{j=1}^n i_j + \nonumber\\
    % + & \prod_{i=1}^n \sum_{j=1}^n j_i
\end{align}%
Therefore,
\begin{align}
\theta^{i-1}_t&=\bar{\theta}-\gamma\sum_{j=0}^{i-2}(1-\alpha^{i-1-j})\frac{\partial \mathcal{L}}{\partial \theta^{j}_s}
\end{align}%
As we tend to derive the form of back-propagation as follows:
\begin{align}
\theta^i_t&=\theta^{i-1}_t - \nabla
\end{align}
Thus,
\begin{align}
&\nabla = \theta^{i-1}_t - \theta^i_t \nonumber\\
       &= Equation 4 - Equation 3 \nonumber\\
       &= \gamma\sum_{j=0}^{i-1}(1-\alpha^{i-j})\frac{\partial \mathcal{L}}{\partial \theta^{j}_s} - \gamma\sum_{j=0}^{i-2}(1-\alpha^{i-1-j})\frac{\partial \mathcal{L}}{\partial \theta^{j}_s} \nonumber\\
       &= \gamma\sum_{j=0}^{i-1}\alpha^{i-1-j}(1-\alpha)\frac{\partial \mathcal{L}}{\partial \theta^{j}_s} \nonumber\\
       &= \gamma(1-\alpha)\sum_{j=0}^{i-1}\alpha^{i-1-j}\frac{\partial \mathcal{L}}{\partial \theta^{j}_s}
\end{align}
In the end, we get the back-propagation formula of EMA based on Equation 13 and 14, denoted as:

\begin{align}
\theta^i_t&=\theta^{i-1}_t - \gamma(1-\alpha)\sum_{j=0}^{i-1}\alpha^{i-1-j}\frac{\partial \mathcal{L}}{\partial \theta^{j}_s}
\end{align}
where $\gamma$ is the learning rate and $(1-\alpha)$ is a small number because $\alpha$ is generally assigned a value close to 1 (e.g., 0.995). Therefore, the optimization of EMA is equivalent to multiplying a small coefficient on the weighted sum of student’s past gradients. With this conservative and ensemble property, the application of EMA can contribute to a more reliable and robust model.

We adopt EMA in SCDL based on the following reasons: (1) The teacher model updated with EMA can quantify the fluctuation of label noise and contributes to consistency predictions. (2) As we justify above, EMA contributes to unbiased predictions with the conservative and ensemble property. (3) EMA doesn’t need back-propagation (BP), which reduces the computational overhead, because BP needs to build the computation graph to compute the gradient.

\subsection{Statistics of Datasets}
The detailed statistics of five publicly available NER datasets are shown in Table 5.
\begin{table}[thp]
\renewcommand\arraystretch{1.1}
\centering
\scriptsize
% \resizebox{6.8cm}{2.2cm}{
\begin{tabular}{cccccc}
\toprule
\multicolumn{2}{c}{\textbf{Dataset}} & Train  & Dev  & Test  & Types \\ \midrule
\multirow{2}{*}{\textbf{CoNLL03}} & \textbf{Sentence}     & 14041      & 3250   & 3453  & \multirow{2}{*}{4}  \\
& \textbf{Token}            & 203621  & 51362  & 46435 \\ \hline
\multirow{2}{*}{\textbf{OntoNotes5.0}} & \textbf{Sentence}     & 115812     & 15680  & 12217  & \multirow{2}{*}{18}  \\
& \textbf{Token}            & 2200865 & 304701 & 230118 \\ \hline
\multirow{2}{*}{\textbf{Webpage}} & \textbf{Sentence}     & 385     & 99  & 135  & \multirow{2}{*}{4}  \\
& \textbf{Token}            & 5293 & 1121 & 1131 \\ \hline
\multirow{2}{*}{\textbf{Wikigold}} & \textbf{Sentence}     & 1142     & 280  & 274  & \multirow{2}{*}{4}  \\
& \textbf{Token}            & 25819 & 6650 & 6538 \\ \hline
\multirow{2}{*}{\textbf{Twitter}} & \textbf{Sentence}     & 2393     & 999  & 3844  & \multirow{2}{*}{10}  \\
& \textbf{Token}            & 44076 & 15262 & 58064 \\ \bottomrule
\end{tabular}%}
\caption*{Table 5: The statistics of datasets.}
\label{appendix Table 1}
\end{table}

% \begin{table*}[b]
% \renewcommand\arraystretch{1.2}
% \centering
% \resizebox{16cm}{1.66cm}{
% \small
% \begin{tabular}{ll}
% \toprule
% \textbf{Noisy Labels 1}  & To evaluate the influence of each component in our method To evaluate the influence of each component in our method\\ \hline
% \textbf{Student $\theta_{s_1}$}  & To evaluate the influence of each component in our method To evaluate the influence of each component in our method\\ \hline
% \textbf{Student $\theta_{t_1}$}  & To evaluate the influence of each component in our method To evaluate the influence of each component in our method\\ \midrule \midrule
% \textbf{Noisy Labels 2} & To evaluate the influence of each component in our method To evaluate the influence of each component in our method\\ \hline
% \textbf{Student $\theta_{s_2}$} & To evaluate the influence of each component in our method To evaluate the influence of each component in our method \\ \hline
% \textbf{Student $\theta_{t_2}$} & To evaluate the influence of each component in our method To evaluate the influence of each component in our method \\ \bottomrule 
% \end{tabular}}
% \caption*{Table 3: Case studies.}
% \label{tab:booktabs}
% \end{table*}
\begin{table}[tb]
\centering
\tiny
% \resizebox{7.7cm}{1.7cm}{
% \begin{adjustbox}{angle=90}
\begin{tabular}{cccccc}
\toprule
\textbf{Hyper Param.} & \textbf{CoNLL03} & \textbf{ON5.0} & \textbf{Webpage} & \textbf{Wikigold} & \textbf{Twitter} \\ \midrule
\textbf{Batch}            & 16 & 32 & 16 & 16 & 16\\ \midrule
\textbf{Epoch}            & 50 & 50 & 50 & 50 & 50\\ \midrule
\textbf{LR}            & 1e-5 & 2e-5 & 1e-5 & 1e-5 & 2e-5\\ \midrule
\textbf{Sche. Warmup}            & 200 & 500 & 100 & 200 & 200\\ \midrule
\textbf{Pre. Epoch}            & 1 & 2 & 12 & 5 & 6\\ \midrule
\textbf{\makecell[c]{Update Cycle\\ (iterations)}}            & 6000 & 7240 & 300 & 2000 & 3200\\ \midrule
\textbf{EMA $\alpha$}            & 0.995 & 0.995 & 0.99 & 0.99 & 0.995\\ \midrule
\textbf{\makecell[c]{Confidence\\ Threshold $\delta$}}            & 0.9 & 0.9 & 0.9 & 0.9 & 0.9\\ \bottomrule
\end{tabular}%}
% \end{adjustbox}
\caption*{Table 6: Hyper-parameter settings.}
\label{appendix Table 2}
\end{table}

\subsection{Hyper-parameter and Baseline Settings}
Detailed hyper-parameter settings for each dataset are shown in Table 6. Specifically, we firstly tune the partial hyper-parameters with Grid-Search for student models (i.e., $\theta_{s_1}$ and $\theta_{s_2}$) (e.g., learning rate chosen from \{1e-5, 2e-5, 5e-5, 1e-4\}, training epoch from \{20, 50, 100\}, batch size from \{16, 32\}). Pretraining epoch is determined when the F1 score on development dataset doesn't increase. The number of steps for the scheduler warmup is chosen from \{100, 200, 500\}. Then we tune EMA $\alpha$ from \{0.9, 0.99, 0.995, 0.998\} for teacher models (i.e., $\theta_{t_1}$ and $\theta_{t_2}$). Finally, we tune update cycle range from 100 to 8000 according to the size of dataset. The confidence threshold is set to 0.9. The rest parameters are default in huggingface Transformers\footnote{https://huggingface.co/transformers/}.

For fair comparison, NegSampling and BOND adopt RoBERTa as the basic model. Co-teaching+ and JoCoR adopt RoBERTa, DistilRoBERTa as the basic models. For NegSampling, we run the officially released code using suggested hyperparameters in the original paper. For Co-teaching+ and JoCoR, noise rate $\tau$ is calculated by distantly supervised and original training set.

We conduct the experiments on NVIDIA Tesla T4 GPU. It is worth noting that only the best model $\theta\in\{ \theta_{t_1}, \theta_{s_1}, \theta_{t_2}, \theta_{s_2}\}$ is adopted for predicting in our SCDL framework. Therefore, the complexity of our model is not increased during the test stage.

\end{document}